\documentclass[acmtog,nonacm]{acmart}
\acmSubmissionID{980}

\usepackage{color}
\usepackage{hyperref}
\usepackage{makecell}
\usepackage{multirow}
\usepackage[caption=false]{subfig}
\usepackage{url}
\usepackage{amsfonts,bm}

\def\eqref#1{equation~\ref{#1}}
\def\1{\bm{1}}

\def\va{{\bm{a}}}

\def\vh{{\bm{h}}}

\def\vs{{\bm{s}}}

\def\vx{{\bm{x}}}

\def\mA{{\bm{A}}}

\def\mM{{\bm{M}}}

\def\mX{{\bm{X}}}

\DeclareMathAlphabet{\mathsfit}{\encodingdefault}{\sfdefault}{m}{sl}
\SetMathAlphabet{\mathsfit}{bold}{\encodingdefault}{\sfdefault}{bx}{n}

\usepackage[ruled]{algorithm2e} % For algorithms

\SetAlFnt{\small}
\SetAlCapFnt{\small}
\SetAlCapNameFnt{\small}
\SetAlCapHSkip{0pt}

\acmJournal{TOG}
\acmVolume{43}
\acmNumber{4}
\acmYear{2024}
\acmMonth{7}

\begin{document}
% Title portion
\title{DiffPoseTalk: Speech-Driven Stylistic 3D Facial Animation and Head Pose Generation via Diffusion Models}

% DO NOT ENTER AUTHOR INFORMATION FOR ANONYMOUS TECHNICAL PAPER SUBMISSIONS TO SIGGRAPH 2019!
\author{Zhiyao Sun}
\orcid{0000-0002-6377-7103}
\author{Tian Lv}
\orcid{0000-0002-3726-0530}
\author{Sheng Ye}
\orcid{0000-0001-9280-6279}
\author{Matthieu Lin}
\orcid{0009-0004-4265-6830}
\author{Jenny Sheng}
\orcid{0009-0007-7863-8409}
\affiliation{%
 \institution{BNRist, Tsinghua University}
 \city{Beijing}
 \postcode{100084}
 \country{China}}
\email{sunzy21@mails.tsinghua.edu.cn}
\email{lt22@mails.tsinghua.edu.cn}
\email{yec22@mails.tsinghua.edu.cn}
\email{yh-lin21@mails.tsinghua.edu.cn}
\email{cqq22@mails.tsinghua.edu.cn}

\author{Yu-Hui Wen}
\authornote{Corresponding authors.}
\orcid{0000-0001-6195-9782}
\affiliation{%
 \institution{Beijing Jiaotong University}
 \city{Beijing}
 \postcode{100044}
 \country{China}}
\email{yhwen1@bjtu.edu.cn}

\author{Minjing Yu}
\orcid{0000-0001-6755-2027}
\affiliation{%
 \institution{Tianjin University}
 \city{Tianjin}
 \postcode{300072}
 \country{China}}
\email{minjingyu@tju.edu.cn}

\author{Yong-Jin Liu}
\authornotemark[1]
\orcid{0000-0001-5774-1916}
\affiliation{%
 \institution{BNRist, Tsinghua University}
 \city{Beijing}
 \postcode{100084}
 \country{China}}
\email{liuyongjin@tsinghua.edu.cn}

\definecolor{Red}{cmyk}{0,1,1,0}
\definecolor{Green}{cmyk}{1,0,1,0}
\definecolor{Cyan}{cmyk}{1,0,0,0}
\definecolor{Purple}{cmyk}{0.45,0.86,0,0}
\definecolor{Rosolic}{cmyk}{0.00,1.00,0.50,0}
\definecolor{Blue}{cmyk}{1.00,1.00,0.00,0}
\definecolor{Orange}{cmyk}{0,0.52,0.80,0}
\definecolor{Black}{cmyk}{1,0,0,1}

\begin{abstract}
The generation of stylistic 3D facial animations driven by speech presents a significant challenge as it requires learning a many-to-many mapping between speech, style, and the corresponding natural facial motion. However, existing methods either employ a deterministic model for speech-to-motion mapping or encode the style using a one-hot encoding scheme. Notably, the one-hot encoding approach fails to capture the complexity of the style and thus limits generalization ability. In this paper, we propose DiffPoseTalk, a generative framework based on the diffusion model combined with a style encoder that extracts style embeddings from short reference videos. During inference, we employ classifier-free guidance to guide the generation process based on the speech and style. In particular, our style includes the generation of head poses, thereby enhancing user perception.
Additionally, we address the shortage of scanned 3D talking face data by training our model on reconstructed 3DMM parameters from a high-quality, in-the-wild audio-visual dataset. Extensive experiments and user study demonstrate that our approach outperforms state-of-the-art methods. The code and dataset are at \url{https://diffposetalk.github.io}.
\end{abstract}




\begin{teaserfigure}
  \setlength{\abovecaptionskip}{3pt}  
  \setlength{\belowcaptionskip}{0pt}
  \centering
  \includegraphics[width=\textwidth]{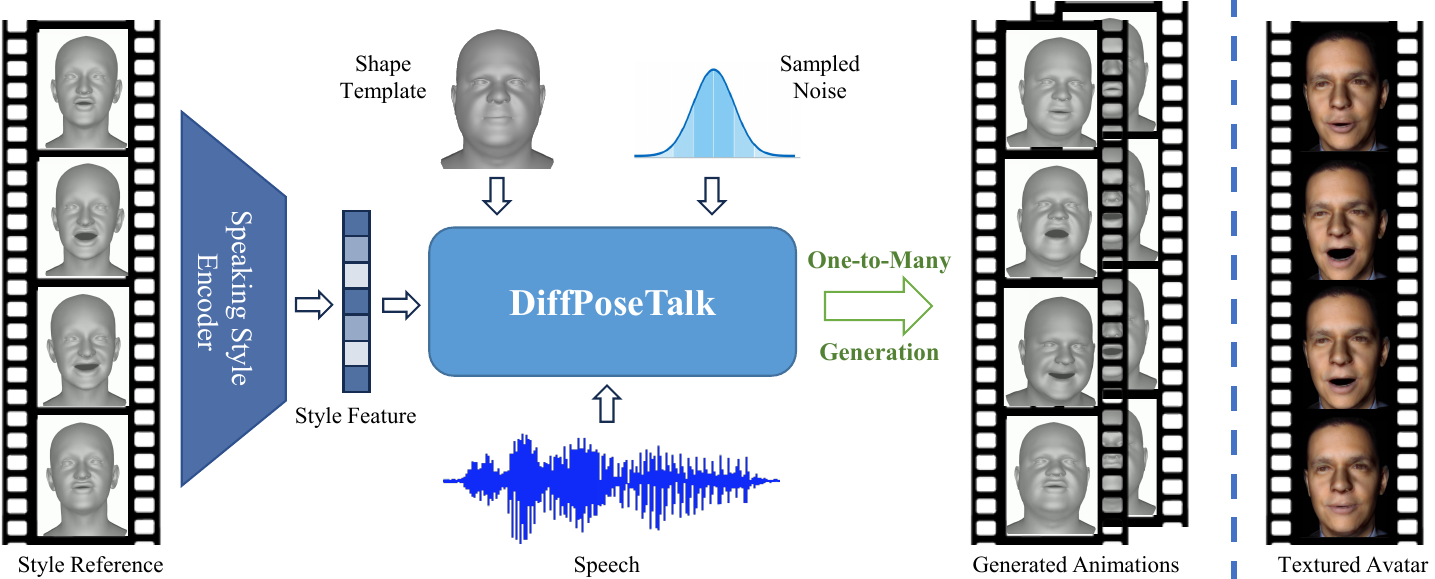}
  \caption{
  We present DiffPoseTalk, a novel diffusion-based speech-driven animation system incorporated with a speaking style encoder to extract style features from arbitrary reference videos. Given an input speech and a speaking style, our system generates diverse and stylistic facial animations along with head movements.
  }
  \label{fig:teaser}
  \Description{Overview of the DiffPoseTalk system.}
  \vspace{0.2cm}
\end{teaserfigure}

\maketitle

\section{Introduction}

The domain of speech-driven 3D facial animation has experienced significant growth in both academia and industry, primarily owing to its diverse applications in education, entertainment, and virtual reality. Speech-driven 3D facial animation generates lip-synchronized facial expressions from an arbitrary speech signal. It is a highly challenging research problem due to the cross-modal many-to-many mapping between the speech and the 3D facial animation. However, most existing speech-driven 3D facial animation methods rely on deterministic models~\citep{VOCA, MeshTalk, FaceFormer, CodeTalker}, which often fail to sufficiently capture the complex many-to-many relationships and suffer from the regression-to-mean problem, thereby resulting in over-smoothed face motions. Furthermore, these methods generally employ a one-hot encoding scheme for representing speaking styles during training, thus limiting their adaptability to new speaking styles.

In contrast to deterministic models, diffusion models can fit various forms of distributions, making them better suited to addressing the many-to-many mapping challenge. Recent diffusion models have shown impressive results in various domains~\citep{DM-survey}. Specifically, the existing diffusion-based audio-driven human motion generation methods have shown appealing results. However, they are not trivially transferable to speech-driven facial animation for three main reasons. First, unlike gestures, which can have a more relaxed temporal correlation with audio (occurring slightly before or after the associated speech), facial movements --- particularly lip motions --- require much stricter timing. This calls for specifically designed structures to precisely align speech and motion information. Second, lip motions contain richer semantics than gestures or dancing, which necessitate a more robust speech encoder to extract phoneme-level features. Lastly, humans have diverse speaking styles. A strong style encoder should be designed to extract style representation from an arbitrary style clip.

To address the aforementioned limitations and challenges, we introduce DiffPoseTalk, a novel controllable diffusion-based generative model, to generate high-quality, diverse, speech-matched, and stylistic facial motions for speech-driven 3D facial animation~(Figure~\ref{fig:teaser}). Our method overcomes the inability of existing diffusion models that cannot be directly transferred to speech-driven expression animation. Compared to existing methods, the main improvement of DiffPoseTalk can be characterized as follows. We use an attention-based architecture to align facial motions with speech, and train a diffusion model to predict the facial expression signal itself~\citep{ramesh2022hierarchical, MDM} instead of predicting the noise; this architecture allows us to facilitate the subsequent addition of geometric constraints to generate more reasonable results. Along with the expression, we also predict the speaker's head pose and design the corresponding loss function to obtain more natural animations. Furthermore, we exploit HuBERT~\citep{HuBERT} to encode the input speech to improve generalization and robustness. Finally, we develop a style encoder to obtain latent style code from a style video clip, and perform classifier-free guidance~\citep{CFG} at inference time to achieve example-based style control. To address the scarcity of co-speech 3D facial animation data by motion capture, we collect and build a speech-driven facial animation dataset with varied speaking styles and head poses.

In summary, our contribution is threefold:

\begin{itemize}
  \item We propose a novel diffusion-based approach to jointly generate diverse and stylistic 3D facial motions with head poses from speech.
  \item We develop a style encoder to extract personalized speaking styles from reference videos, which can be used to guide the motion generation in a classifier-free manner.
  \item We build a new audio-visual dataset that encompasses a diverse range of identities, speaking styles, and head poses. This dataset and our code are available for research purposes.
\end{itemize}

\section{Related Work}

\subsection{Speech-Driven 3D Facial Animation}
% \textbf{Speech-Driven 3D Facial Animation.}
Existing speech-driven 3D facial animation methods can be roughly divided into procedural and learning-based methods. Procedural approaches generally segment speech into phonemes, which are then mapped to predefined visemes via a set of comprehensive rules. For example, \citet{AnimatedSpeech} use dominance functions to map phonemes to corresponding facial animation parameters, while \citet{JALI} factor speech-related animations into jaw and lip actions, employing a co-articulation model to animate facial rigs. Although these procedural methods offer explicit control over the resulting animations, they often require intricate parameter tuning and lack the ability to capture the diversity of real-world speaking styles.

Meanwhile, learning-based methods have grown rapidly over the recent decade. These approaches typically adopt acoustic features like MFCC or pretrained speech model features \citep{DeepSpeech, Wav2Vec2, HuBERT} as the speech representation, which is then mapped to 3D morphable model parameters \citep{Zhang23Pose3DTalk, EmoTalk} or 3D mesh \citep{VOCA, MeshTalk, FaceFormer, CodeTalker, FaceXHuBERT} through neural networks. However, most current methods are regression models, which are deterministic and tend to generate over-smoothed lip motion. This issue is especially pronounced in large-scale datasets, where these methods are prone to yielding average outcomes, thereby struggling to produce precise and diverse responses to voice data. 
Typically, these deterministic approaches have been trained on smaller datasets like VOCA \citep{VOCA} and BIWI \citep{biwi}, which unintentionally sidestep the regression-to-mean challenge. However, when these methods are applied to the more extensive dataset used in our study, a significant decline in performance is observed. In contrast, our method proposed in this paper leverages the strong probabilistic modeling capability of diffusion models to generate diverse and stylistic 3D facial animations. 

Current learning-based methods typically achieve style control in a label-based or example-based manner. The former class relies on manually predefined style labels. For example, \citet{VOCA} and \citet{FaceFormer} employ one-hot embeddings as the style label for different identities in the training set. However, this limits the model's ability to adapt to new individuals and capture complex fine-grained styles. The latter class (e.g., Imitator \citep{Imitator}) is able to generate talking animation in arbitrary styles, even those unseen during training, by imitating examples. Our method falls into this category. Different from Imitator, our method adopts contrastive learning to extract salient style features and does not require optimization or fine-tuning.

\subsection{Diffusion Probabilistic Models}
% \noindent\textbf{Diffusion Probabilistic Models.}
Diffusion probabilistic models~\citep{Sohl-Dickstein15DM, DDPM}, which are able to generate high-quality and diverse samples, have achieved remarkable results in various generative tasks \citep{DM-survey}. They leverage a stochastic diffusion process to gradually add noise to data samples, subsequently employing neural architectures to reverse the process and denoise the samples. A key strength of diffusion models lies in their ability to model various forms of distributions and capture complex many-to-many relationships, making them particularly well-suited for our speech-driven facial animation task.

\begin{figure*}[ht]
\centering
\includegraphics[width=1\linewidth]{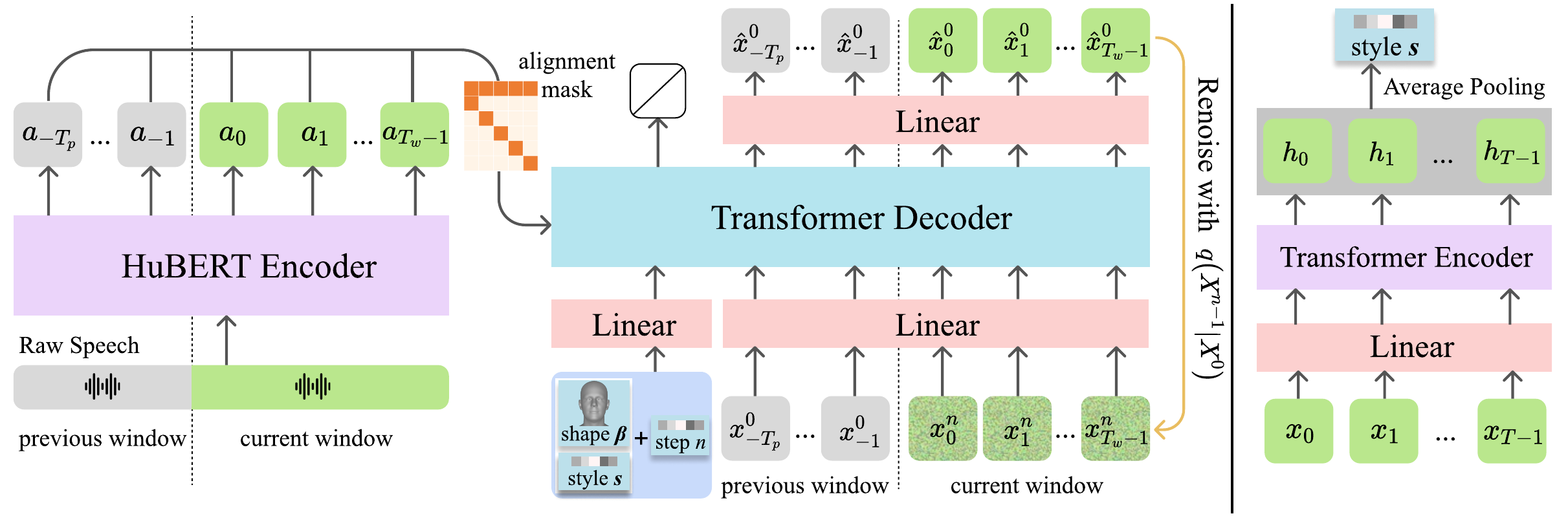}
\caption{\textbf{(Left) Transformer-based denoising network.} We employ a windowing strategy to generate speech-driven 3D facial animations for inputs of arbitrary length. HuBERT-encoded speech features $\mA_{-T_p:T_w}$, prior clean motion parameters $\mX_{-T_p:0}^{0}$, current noisy motion parameters $\mX_{0:T_w}^{n}$, shape parameters $\bm \beta$, style feature $\vs$, and the diffusion timestep $n$ are fed into the transformer decoder. The decoder then predicts clean motion parameters $\hat \mX_{-T_p:T_w}^{0}$, which are renoised to $\mX_{0:T_w}^{n-1}$ for the next denoising iteration. \textbf{(Right) The speaking style encoder.} The style feature $\vs$ can be extracted from a sequence of motion parameters $\mX_{0:T}$ using a transformer encoder.} \label{fig:pipeline}
\Description{The structure of the Transformer-based denoising network and the speaking style encoder.}
\end{figure*}

For conditional generation, classifier-guidance \citep{BeatGAN} and classifier-free guidance \citep{CFG} are widely employed in tasks such as text-to-image synthesis \citep{LatentDiffusion}, text-to-motion \cite{MDM} synthesis, and audio-driven body motion generation \citep{DiffGesture, LDA}. In particular, diffusion model with classifier-free guidance has achieved impressive results in multi-modal modeling. Recently, diffusion models have also been applied in speech-driven 3D facial animation. FaceDiffuser~\citep{facediffuser} leverages the non-deterministic capabilities of diffusion models to capture the complex many-to-many relationship between audio and face motions. Nevertheless, it lacks the ability to control the head poses of the generated talking faces and does not support novel style conditions. In this paper, we propose a novel diffusion-based model that jointly generates facial motions and head poses while accommodating arbitrary novel style conditions.

\section{Method}

An overview of our proposed method is illustrated in Figure~\ref{fig:pipeline}. 
We adopt a well-established, pretrained encoder to extract speech features, while using 3DMM as the face representation (Section~\ref{subsec:3d_face_repr}). A transformer-based denoising network is used for the reverse diffusion process (Section~\ref{subsec:denoise_net}), where we guide the conditional generation in a classifier-free manner (Section~\ref{subsubsec:style_control}).

\subsection{Problem Formulation}

Our method takes a speech feature\footnote{We use Python style indexing and slicing in this paper, \textit{i. e.}, ``$0:T$'' includes ``$0, 1, \dots, T-1$''.} $\mA_{0:T}$, the 3DMM shape parameter $\bm{\beta}$ of a
template face, and a speaking style vector $\vs$ as input, and generates a 3DMM-based 3D facial animation represented by a sequence of 3DMM expression and pose parameters $\mX_{0:T}$. The style vector $\vs$ can be extracted from a short reference video using our speaking style encoder (Section~\ref{subsec:style_encoder}). 

\noindent\textbf{Speech Representation.} \label{subsec:speech_repr}
Extensive facial animation studies have shown that self-supervised pretrained speech model features like Wav2Vec2 \citep{Wav2Vec2} and HuBERT \citep{HuBERT} outperform traditional ones such as MFCC. Based on these findings, we utilize HuBERT as our chosen speech encoder for facial animation generation, as it has been proven to be superior to Wav2Vec2 in this regard \cite{FaceXHuBERT}.
HuBERT consists of a temporal convolutional audio feature extractor and a multi-layer transformer encoder. To align the audio features with the facial motions' sampling rate, we introduce a resampling layer after the temporal convolutions. Thus, for a given raw audio clip that matches a facial motion sequence of length $T$, HuBERT generates a speech representation $\mA_{0:T}$ that also spans $T$ time steps.

\noindent\textbf{3D Face Representation.} \label{subsec:3d_face_repr} We use a 3D morphable model FLAME~\citep{FLAME} with $N=5,023$ vertices and $K=4$ joints, whose geometry can be represented using parameters $\{\bm \beta, \bm \psi, \bm \theta\}$, where $\bm \beta \in \mathbb R^{100}$ is the shape parameter, $\bm \psi \in \mathbb R^{50}$ is the expression parameter, and $\bm \theta \in \mathbb R^{3K+3}$ is the head pose parameter. Given a set of FLAME parameters, the 3D face mesh can be obtained with $
    M(\bm \beta, \bm \theta, \bm \psi) = W\left(T_P(\bm \beta, \bm \theta, \bm \psi), \mathbf J(\bm \beta), \bm \theta, \mathcal W\right),
$
where $T_P$ outputs vertices by combining blend shapes, the standard skinning function $W(\mathbf T, \mathbf J, \bm \theta, \mathcal W)$ rotates the vertices of $\mathbf T$ around joints $\mathbf J$, and $\mathcal W$ performs linear smoothing.
Specifically, we use the shape parameter $\bm \beta$ to serve as the neutral template face for the speaker. For facial animation, we predict the expression parameters $\bm \psi$ as well as the jaw and global rotation components within the pose parameters $\bm \theta$. To simplify notation, we compose $\bm \psi$ and $\bm \theta$ as the {\it motion parameter} $\vx$ and rewrite the mesh construction function as $M(\bm \beta, \vx)$. The reasons for choosing 3DMM parameters over mesh vertices as the face representation are discussed in Section~\ref{subsec:choice_of_face}.

To reconstruct accurate 3DMM parameters from the audio-visual dataset, we make comprehensive use of several state-of-the-art 3D face reconstruction and pose estimation works, similar to \citet{danvevcek2023emotional}. We adopt MICA \citep{MICA} for identity shape prediction, SPECTRE \citep{SPECTRE} for accurate expression reconstruction of lip movements and jaw pose prediction, and 6DRepNet \citep{6DRepNet} for head pose prediction. Then, we follow EmoTalk \citep{EmoTalk} to apply a Savitzky-Golay filter to the predicted expressions and poses, which markedly improves motion smoothness.

\subsection{Facial Animation Diffusion Model} \label{subsec:denoise_net}

We propose to use a diffusion model to generate speech-driven facial animation. The diffusion model involves two processes. The forward process is a Markov chain $q\left(\mX^{n} | \mX^{n-1} \right)$ for $n \in \{1, \dots, N\}$ that progressively adds Gaussian noise to an initial data sample $\mX^{0}$ according to a variance schedule. The original sample is gradually substituted by noises, eventually reaching a standard normal distribution $q\left(\mX^{N} | \mX^{0} \right)$. The reverse process, on the contrary, leverages the distribution $q\left(\mX^{n-1} | \mX^{n} \right)$ to recreate the sample from noise. This distribution depends on the entire dataset and hence is intractable. Therefore, a denoising network is used to approximate this distribution. In practice, the denoising network is trained to predict the noise \citep{DDPM} or the clean sample $\mX^{0}$ \citep{DM-Sample}. We opt for the latter, as it enables us to incorporate geometric losses that offer more precise constraints on facial motions. The effectiveness of this scheme has been validated by prior works on human body motion generation \citep{MDM} and our experiments.

\noindent\textbf{Transformer-Based Denoising Network.} Our transformer-based denoising network, as illustrated in Figure~\ref{fig:pipeline}, consists of two components: a pretrained HuBERT encoder for extracting speech features $\mA$, and a transformer decoder for sampling predicted motions $\hat \mX^0$ from noisy observations $\mX^n$ ($n = N, N - 1, \dots, 1$) in an iterative manner. A notable design is an alignment mask between the encoder and the decoder~\cite{FaceFormer}, which ensures proper alignment of the speech and motion modalities. Specifically, the motion feature at position $t$ only attends to the speech feature $\va_t$. The initial token, which is composed of diffusion timestep $n$ and other conditions, attends to all speech features. We allow the transformer part of the HuBERT speech encoder to be trainable, which enables HuBERT to better capture motion information directly from speech. To accommodate sequences of arbitrary lengths, we implement a windowing strategy for the inputs. 

Formally, the inputs to the denoising network are processed as follows. For a given speech feature sequence of length $T$, we partition it into windows of length $T_w$ (padding is added to the audio if it is not long enough). To ensure seamless transitions between consecutive windows, we include the last $T_p$ frames of speech features $\mA_{-T_p:0}$ and motion parameters $\mX_{-T_p:0}^{0}$ from the preceding window as conditional inputs. 
Note that for the first window, the speech features and motion parameters are replaced with learnable start features $\mA_\text{start}$ and $\mX_\text{start}$. Within each window at diffusion step $n$, the network receives both previous and current speech features $\mA_{-T_p:T_w}$, the previous motion parameters $\mX_{-T_p:0}^{0}$, and the current noisy motion parameters $\mX_{0:T_w}^{n}$ sampled from $q\left(\mX_{0:T_w}^{n} | \mX_{0:T_w}^{0} \right)$. The denoising network then outputs the clean sample as:
\begin{equation}
    \hat \mX_{-T_p:T_w}^{0} = D\left(\mX_{0:T_w}^{n}, \mX_{-T_p:0}^{0}, \mA_{-T_p:T_w}, n \right).
\end{equation}

% \subsubsection{Losses}
\noindent\textbf{Losses.} We use the simple loss \citep{DDPM} for the predicted sample:
\begin{equation}
    \mathcal L_\text{simple} = \left\| \hat \mX_{-T_p:T_w}^{0} - \mX_{-T_p:T_w}^{0} \right\|^2.
\end{equation}
To better constrain the generated face motion, we convert the FLAME parameters into zero-head-posed 3D mesh sequences. Formally, $\mM_{-T_p:T_w} = M_0\left(\bm \beta, \mX_{-T_p:T_w}^{0} \right)$  and $\hat \mM_{-T_p:T_w} = M_0\left(\bm \beta, \hat \mX_{-T_p:T_w}^{0} \right)$. We then apply the following geometric losses in 3D space: the vertex loss $\mathcal L_\text{vert}$ \citep{VOCA} for the positions of the mesh vertices, the velocity loss $\mathcal L_\text{vel}$ \citep{VOCA} for better temporal consistency, and a smooth loss $\mathcal L_\text{smooth}$ to penalize large acceleration of the predicted vertices:
\begin{equation} \label{eq:vert_loss}
    \mathcal L_\text{vert} = \left\| \mM_{-T_p:T_w} - \hat \mM_{-T_p:T_w} \right\|^2,
\end{equation}
\begin{equation} \label{eq:velocity_loss}
    \mathcal L_\text{vel} = \left\| \left(\mM_{-T_p+1:T_w} - \mM_{-T_p:T_w-1} \right) - \left(\hat \mM_{-T_p+1:T_w} - \hat \mM_{-T_p:T_w-1} \right) \right\|^2,
\end{equation}
\begin{equation} \label{eq:smooth_loss}
    \mathcal L_\text{smooth} = \left\| \hat \mM_{-T_p+2:T_w} - 2 \hat \mM_{-T_p+1:T_w-1} + \hat \mM_{-T_p:T_w-2} \right\|^2.
\end{equation}
We apply geometric losses $\mathcal L_\text{head}$ to head motions in a similar way. Please refer to the Appendix %Section~\ref{ap:subsec:head_loss} 
for more details.

In summary, our overall loss is defined as:
\begin{equation} \label{eq:loss}
    \mathcal L =  \mathcal L_\text{simple} + \lambda_\text{vert} \mathcal L_\text{vert} + \lambda_\text{vel} \mathcal L_\text{vel} + \lambda_\text{smooth} \mathcal L_\text{smooth} + \mathcal L_\text{head}.
\end{equation}

\subsection{Style-Controllable Diffusion Model} \label{subsubsec:style_control}

The facial animation diffusion model generates facial motions conditioned on input speech. In addition to the speech, we use speaking style and template face shape as control conditions.
The shape parameters $\bm \beta$ and speaking style feature $\vs$ are shared across all windows. The denoising network then outputs the clean sample as:
\begin{equation}
    \hat \mX_{-T_p:T_w}^{0} = D\left(\mX_{0:T_w}^{n}, \mX_{-T_p:0}^{0}, \mA_{-T_p:T_w}, \vs, \bm \beta, n \right).
\end{equation}

\subsubsection{Speaking Style Encoder} 
%\noindent\textbf{Speaking Style Encoder.}
\label{subsec:style_encoder} We introduce a novel speaking style encoder designed to capture the unique speaking style of a given speaker from a brief video clip. Speaking style is a multifaceted attribute that manifests in various aspects such as the size of the mouth opening \citep{VOCA}, facial expression dynamics --- especially in the upper face \citep{CodeTalker} --- and head movement patterns \citep{ADTF, Zhang23Pose3DTalk}. Given the complexity and difficulty in quantitatively describing speaking styles, we opt for an implicit learning approach through contrastive learning. We operate under the assumption that the short-term speaking styles of the same person at two proximate times should be similar.

\textit{Architecture.}
The speaking style encoder (Figure~\ref{fig:pipeline} right) utilizes a transformer encoder to extract style features from a sequence of motion parameters $\mX_{0:T}$. The encoder features $\{ \vh_i \}$ are aggregated by average pooling into the style embedding $\vs$. Formally, this is described as: 
\begin{equation}
    \vs = SE\left(\mX_{0:T}\right).
\end{equation}

\textit{Contrastive Learning.}
We use the NT-Xent loss \citep{SimCLR} for contrastive learning. Each training minibatch consists of $N_s$ samples of speech features and motion parameters of length $2T$. We split the sample length in half to get $N_s$ pairs of positive examples. Given a positive pair, the other $2(N_s-1)$ examples are treated as negative examples. We use cosine similarity as the similarity function. The loss function for a positive pair of examples $(i, j)$ is defined as:
\begin{equation}
    \mathcal L_{i, j} = - \log \frac{\exp \left(\mathrm{cos\_sim}(\vs_i, \vs_j) / \tau \right)}{\sum_{k=1}^{2N_s} \1_{k \ne i} \exp \left(\mathrm{cos\_sim}(\vs_i, \vs_k) / \tau \right)},
\end{equation}
where $\1_{k \ne i}$ is an indicator function and $\tau$ represents a temperature parameter. The overall loss is computed across all positive pairs for both $(i, j)$ and $(j, i)$.

\subsubsection{Training Strategy}
%\noindent\textbf{Training Strategy.}
In our window-based generation approach, our network faces two different scenarios: (a) generating the initial window, where the previous window conditions are learnable start features, and (b) generating subsequent windows, where the conditions are speech features and motions parameters from the preceding window. The network also requires a shape parameter $\bm \beta$ and a speaking style feature $\vs$ in both scenarios. Therefore, we propose a novel training strategy to meet this demand. Specifically, each training sample includes a frame of shape parameter $\bm \beta$, a speech clip (which will be encoded into speech features $\mA_{0:2T_w}$ by the HuBERT encoder), and a corresponding motion parameter sequence $\mX_{0:2T_w}^{0}$. We partition the sample into two windows and employ the speaking style encoder to derive style features for each, resulting in $(\vs_a, \vs_b)$. The tuple $(\mA_{0:T_w}, \mX_{0:T_w}^{0}, \vs_b)$ is used to train the first window, while $(\mA_{T_w:2T_w}, \mX_{T_w:2T_w}^{0}, \vs_a)$ with the previous window conditions $(\mA_{T_w-T_p:T_w}, \mX_{T_w-T_p:T_w}^{0})$ is used to train the second window. Taking into account that the actual length of speech during generation may not fully occupy the window, we introduce a random truncation of samples during training. This approach ensures that the model is robust to variations in speech length.

\subsubsection{Sampling with Incremental Classifier-Free Guidance} 
%\noindent\textbf{Sampling with Incremental Classifier-Free Guidance.}
During generation, we sample the result $\mX^{0}$ conditioned on $(\mA, \vs, \bm \beta)$ in an iterative manner. Specifically, we estimate the clean sample as $\hat \mX^{0} = D\left(\mX^{n}, \mA, \vs, \bm \beta, n \right)$ and subsequently reintroduce noise to obtain $\mX^{n-1}$. This process is repeated for $n = N, N-1, \dots, 1$, in accordance with the process in \citet{MDM}. 

Furthermore, we find it beneficial to apply classifier-free guidance \citep{CFG} with the incremental scheme \citep{InstructPix2Pix}, which has been successfully applied to image generation from multiple conditions, where
\begin{equation}
    \begin{split}
    \hat \mX^{0} = D\left(\mX^{n}, \emptyset, \emptyset, \bm \beta, n \right) & + w_a \left[D\left(\mX^{n}, \mA, \emptyset, \bm \beta, n \right) - D\left(\mX^{n}, \emptyset, \emptyset, \bm \beta, n \right) \right] \\
    & + w_s \left[D\left(\mX^{n}, \mA, \vs, \bm \beta, n \right) - D\left(\mX^{n}, \mA, \emptyset, \bm \beta, n \right) \right].
    \end{split}
\end{equation}
The $w_a$ and $w_s$ are the guidance scales for audio and style, respectively. During training, we randomly set the style condition to $\emptyset$ with 0.45 probability, and set both the audio and style conditions to $\emptyset$ with 0.1 probability.

\section{Experiments}

\subsection{Datasets} 

We introduce a new dataset--- Talking Face with Head Poses~(TFHP) --- which contains 1,052 videos of 588 subjects, totaling 26.5 hours. In TFHP, 348 videos are collected from the downloading script provided by High-Definition Talking Face (HDTF) dataset \citep{HDTF}. Compared with HDTF, our TFHP dataset is more diversified in content, featuring video clips from lectures, online courses, interviews, and news programs, thereby capturing a wider array of speaking styles and head movements. Moreover, all videos are converted to 25 fps. In total, approximately 2,385,000 frames of FLAME parameters are reconstructed from the videos with our carefully designed data processing pipeline as previously mentioned. We split the combined dataset by speakers, resulting in 460 for training, 64 for validation, and 64 for testing. 

\begin{table*}
    \renewcommand{\arraystretch}{1.2}
    \centering
    \caption{Quantitative evaluation of the comparative methods, our proposed method, and ablation study variants. We run the evaluation 10 times and report the average score with a $95\%$ confidence interval when applicable. We also report the diversity scores of expression and head pose generation. Note that the vertex-related metrics are not comparable with SadTalker due to its different face topology.}
    \label{tab:metrics}
    \vspace{0.25cm}
    \begin{tabular}{c|c|cccc|cc} 
         \Xhline{2\arrayrulewidth}
         & Methods&  LVE (mm) $\downarrow$ &  FDD ($\times 10^{-5}$m) $\downarrow$ &   MOD (mm) $\downarrow$ & BA $\uparrow$ & Div (exp) ($\times 10^{-4}$) $\uparrow$ & Div (HP) $\uparrow$ \\ 
         \Xhline{2\arrayrulewidth}
         \multirow{4}{*}{\rotatebox[origin=c]{90}{w/o HP}} 
         & FaceFormer&  $9.90^{\pm0.030}$&  $16.95^{\pm0.016}$&   $2.63^{\pm0.015}$&N/A& 0 & N/A \\ 
         & CodeTalker&  $12.71^{\pm0.057}$&  $12.44^{\pm0.064}$&   $2.87^{\pm0.034}$&N/A& 0 & N/A\\ 
         & FaceDiffuser&  $12.12^{\pm0.038}$&  $15.48^{\pm0.048}$&   $3.50^{\pm0.052}$&N/A& $5.93\times10^{-5}$ & N/A\\ 
         & Ours (no HP)& $\mathbf{8.81^{\pm0.008}}$& $10.13^{\pm0.038}$&  $1.72^{\pm0.009}$&N/A& $\mathbf{2.83}$ & N/A\\ 
         \hline
         \multirow{3}{*}{\rotatebox[origin=c]{90}{w/ HP}} 
         & Yi et al.&  $9.99$&  $21.50$&   $2.42$& $0.26$& 0 & 0 \\ 
         & SadTalker&  ---&  ---&   ---&$0.24^{\pm0.001}$& 0 & $0.808$\\ 
         & Ours& $8.94^{\pm0.013}$& $9.60^{\pm0.027}$&  $1.62^{\pm0.009}$&${\mathbf{0.29^{\pm0.005}}}$ & $\mathbf{2.19}$ & $\mathbf{1.16}$\\ 
         \hline
         \multirow{4}{*}{\rotatebox[origin=c]{90}{Ablations}} 
         & Ours w/o $\mathcal L_{geo}$ & $11.29^{\pm0.012}$ & $15.11^{\pm0.043}$ & $2.14^{\pm0.013}$ & $0.28^{\pm0.009}$&  ---&  ---\\
         & Ours w/o AM& $12.81^{\pm0.011}$& $12.58^{\pm0.049}$&  $2.18^{\pm0.007}$&$0.24^{\pm0.006}$&  ---&  ---\\ 
         & Ours w/o CFG & $9.58^{\pm 0.014}$ & $\mathbf{9.59^{\pm0.032}}$ & $\mathbf{1.56^{\pm0.011}}$ & $\mathbf{0.29^{\pm0.010}}$&  ---&  ---\\
         & Ours w/o SSE& $11.33^{\pm0.013}$& $12.97^{\pm0.052}$&$2.03^{\pm 0.017}$ &$0.28^{\pm0.005}$&  ---&  ---\\
         \Xhline{2\arrayrulewidth}
    \end{tabular}
\end{table*}

\subsection{Experiment Setup}
\noindent\textbf{Implementation Details.} We use a four-layer transformer encoder with four attention heads for the speaking style encoder, with feature dimension $d_s = 128$, sequence length $T = 100$ (4 seconds), and temperature $\tau = 0.1$. We train the encoder with the Adam optimizer \citep{Adam} for 26k iterations, with a batch size of 32 and a learning rate of $1\mathrm{e}{-4}$.

We use an eight-layer transformer decoder with eight attention heads for the denoising network, with the feature dimension $d = 512$, the window length $T_w = 100$, and $T_p = 10$. We adopt a cosine noise schedule with diffusion $N = 500$ steps. We train the denoising network with the Adam optimizer for 90k iterations, using 5k warmup steps, batch size 16, and learning rate $1\mathrm{e}{-4}$. We set $\lambda_\text{vert} = 2\mathrm{e}{6}$, $\lambda_\text{vel} = 1\mathrm{e}{7}$, and $\lambda_\text{smooth} = 1\mathrm{e}{5}$ to balance the magnitude of the losses. The overall training on an Nvidia 3090 GPU takes about 12 hours.

%\subsection{Comparison with States of the Arts}
\noindent\textbf{Baselines.}
We compare our approach with state-of-the-art 3D facial animation methods: FaceFormer \citep{FaceFormer}, CodeTalker \citep{CodeTalker}, and FaceDiffuser~\citep{facediffuser}, which are trained with 3D mesh data generated from 3DMM parameters of our TFHP dataset. Recognizing the scarcity of speech-driven 3D animation methods that account for head poses, we also compare with two 2D talking face generation methods, \citet{ADTF} and SadTalker \citep{SadTalker}, which incorporate head movements and utilize a 3DMM as an intermediate face representation. To compare with these two types of methods, we train two versions of our model: ``Ours'' (with head pose prediction) and ``Ours (no HP)'' (without head pose prediction).

\subsection{Quantitative Evaluation}
%\noindent\textbf{Quantitative Evaluation.} 
Following previous studies, we employ two established metrics --- lip vertex error (LVE) \citep{MeshTalk} and upper face dynamics deviation (FDD) \citep{CodeTalker} --- for the quantitative evaluation of generated facial expressions. LVE measures lip synchronization by calculating the maximum L2 error across all lip vertices for each frame. FDD evaluates the upper face motions, which are closely related to speaking styles, by comparing the standard deviation of each upper face vertex's motion over time between the prediction and the ground truth. To assess head motion, we use beat alignment (BA) \citep{Bailando, SadTalker}, albeit with a minor modification: we compute the synchronization of detected head movement beats between the predicted and the actual outcomes. For examining the diversity of facial expressions and head poses generated from \textit{identical} input, we follow \citet{DiffusionMotion} to compute a diversity score. Since the size of the mouth opening can also indicate speaking style \citep{VOCA}, we introduce a new metric called {\it mouth opening difference} (MOD), which measures the average difference in the size of the mouth opening between the prediction and ground truth.

We present the quantitative results and the diversity scores in Table~$\ref{tab:metrics}$. Our method outperforms all others across all metrics, achieving the best lip synchronization and head pose beat alignment. Additionally, the FDD, MOD, and BA metrics suggest that our method most effectively captures speaking styles. As for diversity, the other methods employ deterministic approaches for motion generation, with the exception of SadTalker (which samples head poses from a VAE) and FaceDiffuser. However, we find FaceDiffuser produces nearly identical results for the same input. Consequently, these methods are unable to produce varied expressions and head poses from identical inputs, falling short in capturing this many-to-many mapping.

\begin{figure*}[ht]
\centering
\includegraphics[width=0.99\linewidth]{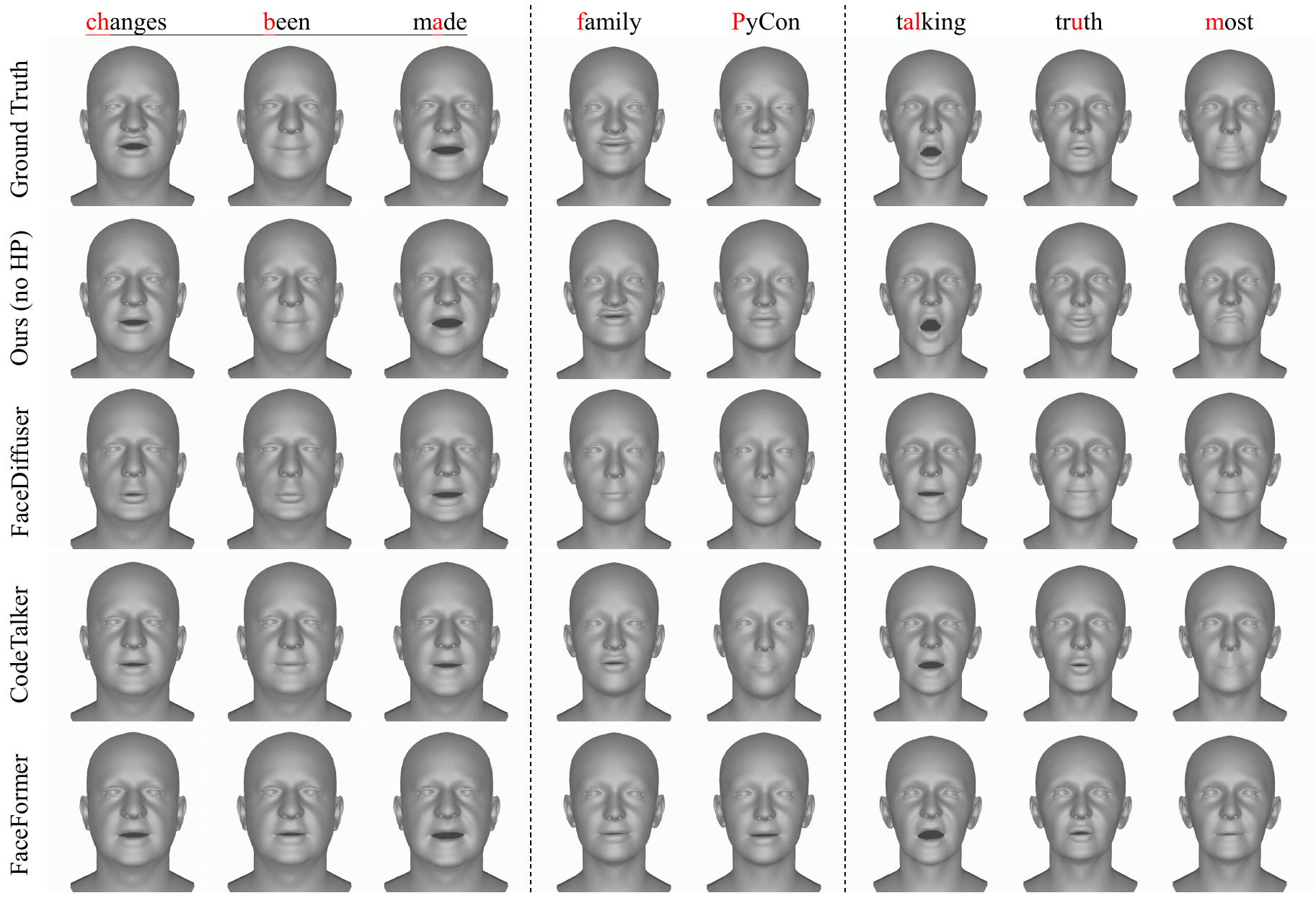}
\caption{Qualitative comparison with the state of the arts (w/o head pose prediction). Results of different identities are split by dashed lines.} \label{fig:comparison1}
\Description{
The figure provides a qualitative comparison with the state of the arts, specifically those without head pose prediction capabilities. It showcases results for different identities and phonemes.
}
\end{figure*}

\subsection{Qualitative Evaluation}
\begin{figure*}[ht]
\centering
\includegraphics[width=0.99\linewidth]{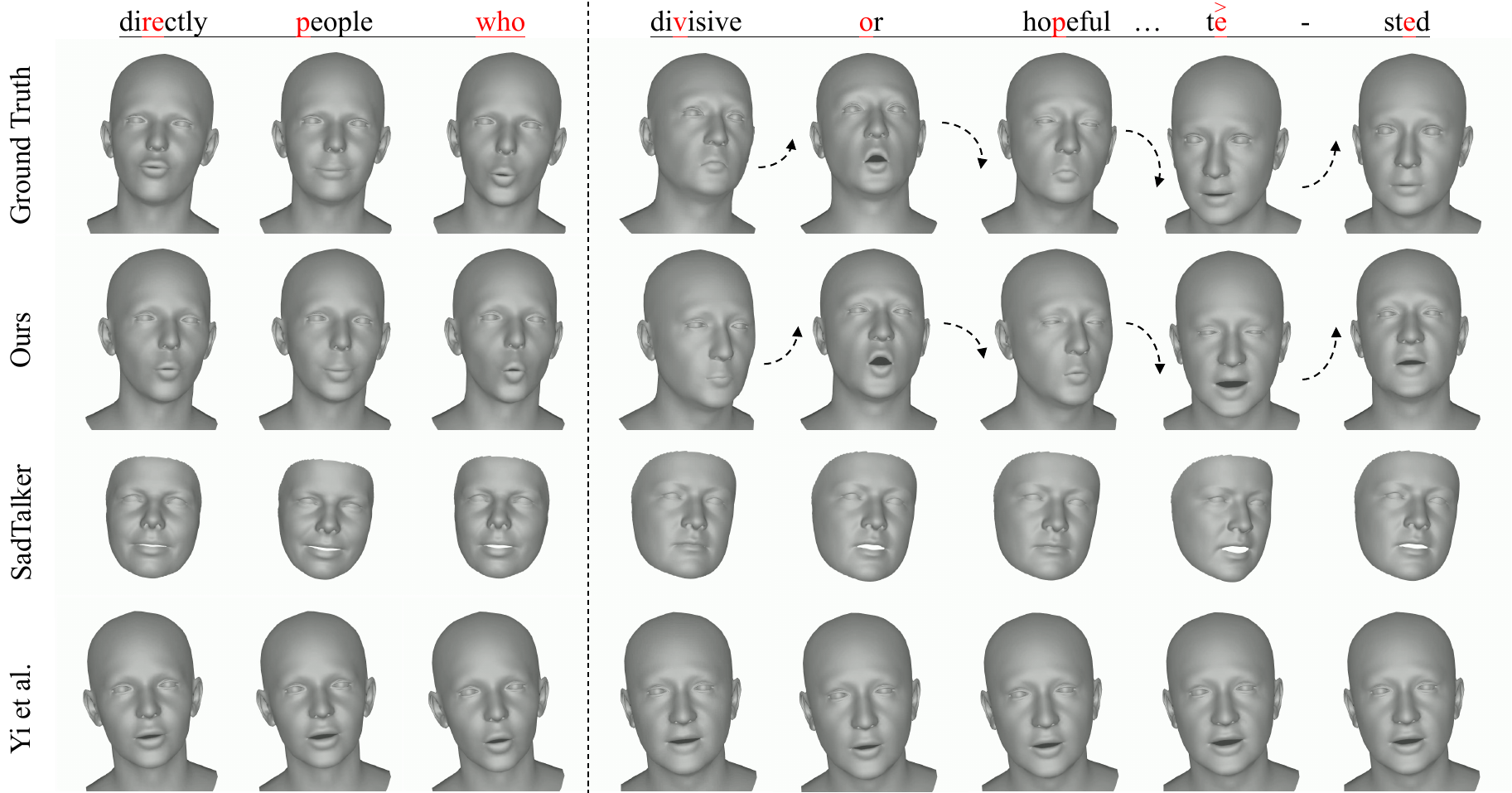}
\caption{Qualitative comparison with the state of the arts (w/ head pose prediction). Results of different identities are split by dashed lines. The ``>'' indicates stress in speech.} \label{fig:comparison2}
\Description{
The figure provides a qualitative comparison with the state of the arts, specifically those with head pose prediction capabilities. It showcases results for different identities and phonemes.
}
\end{figure*}

%\noindent\textbf{Qualitative Evaluation.}
We show the comparison of our method with other comparative methods in Figure~\ref{fig:comparison1}~and~\ref{fig:comparison2}. Our method excels in handling challenging cases, such as articulating bilabial consonants and rounded vowels. Moreover, the generated results have the closest speaking style to the ground truth in aspects like upper facial expressions and mouth opening size. Notably, our approach can also spontaneously produce random movements like eye blinks, which are implicitly learned from the data distribution. Our generated head motions also align well with the stress and rhythm in speech, in a way similar to the ground truth. More results can be found in the supplementary demo video.

\begin{table}
    \renewcommand{\arraystretch}{1.1}
    \caption{User study results.}
    \label{tab:user_study}
    \centering
    \begin{tabular}{cccc}
    \Xhline{2\arrayrulewidth}
    Method & Lip Sync $\uparrow$ & Style Sim $\uparrow$ & Natural $\uparrow$ \\
    \Xhline{2\arrayrulewidth}
    FaceFormer & 2.56 & 2.60 & 2.36 \\
    CodeTalker & 2.88 & 3.00 & 2.90 \\
    FaceDiffuser & 2.71& 2.51 & 2.35 \\
    Ours (no HP) & \bf{4.23} & \bf{4.07} & \bf{4.43}\\
    \hline
    Yi et al. & 1.94 & 2.02 & 1.99 \\
    SadTalker & 3.25 & 2.91 & 2.96 \\
    Ours & \bf{4.52} & \bf{4.25} & \bf{4.43}\\
    \Xhline{2\arrayrulewidth}
    \end{tabular}
\end{table}

\subsection{User Study}
%\noindent\textbf{User Study.} 
To conduct a more comprehensive assessment of our approach, we designed a user study with the following experiment setup. The methods to be compared are categorized into two groups based on whether head motion are generated. The group without head motion includes FaceDiffuser, FaceFormer, CodeTalker, and Ours (no HP) and consists of 12 sets of questions. The group with head motion involves Yi et al., SadTalker, and our approach, comprising 8 sets of questions. In each set, participants are shown the ground truth animation as well as animations generated by each method. Participants are then asked to rate on a scale of 1-5 the lip synchronization, similarity to the speaking style of the ground truth, and the naturalness of facial expressions and head movements. For more information about the settings, please refer to the Appendix. Twenty-six participants took part in the study, and the results are presented in Table~\ref{tab:user_study}. The results demonstrate that our method significantly outperforms existing works in terms of lip synchronization, similarity to the ground truth speaking style, and the naturalness of facial expressions and head movements.

\subsection{Ablation Study}

The results are summarized in Table~\ref{tab:metrics}, showing that the removal of the speaking style encoder (Ours w/o SSE) leads to a decline in performance across all metrics, as our method is no longer capable of generating animations tailored to the speaking style of individuals. Removing all geometric losses (Ours w/o $\mathcal L_{geo}$) results in our method being unable to produce precise facial animations. Removing the alignment mask (Ours w/o AM) causes a serious out-of-sync problem. However, we observe that excluding classifier-free guidance (Ours w/o CFG) yields a mixed impact on the metrics. This is probably due to the fact that CFG is a technique in diffusion models that aims to reduce the diversity of the generated samples while enhancing the quality of each individual sample. Thus, CFG improves LVE for lip synchronization, while slightly diminishing performance on FDD and MOD evaluations, which we speculate are metrics closely related to speaking styles.

\section{Discussions} \label{subsec:choice_of_face}

\textbf{Choice of Face Representation.}
Unlike closely related works \citep{VOCA, FaceFormer, CodeTalker} that use 3D mesh vertices, our approach leverages a widely used 3DMM --- specifically the FLAME model \citep{FLAME} --- for face representation. There are several reasons for this choice. First, given the computational intensity of diffusion models, the lower-dimensional 3DMM parameters offer a substantial advantage in terms of computational speed when compared to predicting mesh vertices. Second, data collection and coverage present challenges. Capturing real-world 3D mesh data requires professional motion capture systems and considerable investments of time and effort, thereby constraining the scale and diversity of data that can be collected. For example, VOCASET \citep{VOCA} only provides less than 30 minutes of data from just 12 subjects. Conversely, 2D audio-visual data are much simpler to collect, and numerous off-the-shelf methods exist for reconstructing 3DMM parameters from these data, offering considerably wider coverage of identities, phonemes, and styles than scanned mesh data. Additionally, the reduced dimensionality, along with the blendshapes as a prior, simplifies the learning process and improves generalization \citep{EmoTalk}. Lastly, using 3DMM parameters facilitates integration with downstream applications such as driving blendshape-based avatars \citep{GaussianAvatars} or facial expression editing \citep{geng20193d, FEE4TV}. We demonstrate such rendered avatar animations in the Appendix and demo video.

\noindent \textbf{Limitations.} Although our method is able to generate high-quality stylistic 3D talking face animation with vivid head poses, there are still some limitations within our framework that could be addressed in follow-up works. Firstly, the computational cost of inference is relatively high due to the sequential nature of the denoising process. To mitigate this, future research can explore more advanced denoisers such as DPM-solver++ \citep{DPM-SolverPP}. Secondly, our method may produce vague or overly smoothed lip motions when encountering very high noise. Possible solutions include incorporating noise suppression during preprocessing or augmenting the training data with noisy audio. Additionally, like existing SOTAs, our approach focuses on animating the face shape while ignoring the inner mouth (including teeth and tongue). Exploring the representation and animation of the inner mouth can lead to more realistic results. Lastly, a promising direction for future research would be to collect real-world 3D talking data that encompasses a broader range of identities and styles, which would further enhance the effectiveness of our approach and contribute to the research community.

\noindent \textbf{Ethical Considerations.} Since our approach is able to generate realistic talking head sequences, there are risks of misusing, such as deepfake generation and deliberate manipulation. Therefore, we firmly require that all talking face sequences generated by our method be marked or noted as synthetic data. Moreover, we will make our code publicly available to the deep fake detection community to further ensure that our proposed method can be applied positively. We also hope to raise awareness of the risks and support regulations preventing technology abuse involving synthetic videos.

\section{Conclusion}

Speech-driven expression animation has a wide range of applications in daily life and has received extensive attention from the research community. It involves a challenging many-to-many mapping across modalities between speech and expression animation. In this paper, we present DiffPoseTalk, a novel diffusion-based method for generating diverse and stylistic 3D facial animations and head poses from speech. We leverage the capability of the diffusion model to effectively replicate the distribution of diverse forms, thereby addressing the many-to-many mapping challenge. Additionally, we resolve the limitations associated with current diffusion models that hinder its direct application to speech-driven expression animation. Leveraging the power of the diffusion model and the speaking style encoder, our approach excels in capturing the many-to-many relationships between speech, style, and motion.

\bibliographystyle{ACM-Reference-Format}
\bibliography{sample-bibliography}

\end{document}